\documentclass[letterpaper,mlapa]{article}
\usepackage{ml2k}
\usepackage{mlapa}

\begin{document}

\twocolumn[
\icmltitle{Bootstrapping Syntax and Recursion\\using Alignment-Based Learning}
\icmlauthor{Menno~van~Zaanen}{menno@scs.leeds.ac.uk}
\icmladdress{School of Computer Studies,
University of Leeds,
Woodhouse Lane,
LS2 9JT Leeds,
UK}
\vskip 0.15in
]

\begin{abstract}
This paper introduces a new type of unsupervised learning algorithm, based on
the alignment of sentences and Harris's \yrcite{bib:misl} notion of
interchangeability. The algorithm is applied to an untagged, unstructured corpus
of natural language sentences, resulting in a labelled, bracketed version of the
corpus. Firstly, the algorithm aligns all sentences in the corpus in pairs,
resulting in a partition of the sentences consisting of parts of the sentences
that are similar in both sentences and parts that are dissimilar. This
information is used to find (possibly overlapping) constituents. Next, the
algorithm selects (non-overlapping) constituents. Several instances of the
algorithm are applied to the ATIS corpus \cite{bib:balacoetpt} and the
OVIS\footnote{Openbaar Vervoer Informatie Systeem (OVIS) stands for Public
Transport Information System.} corpus \cite{bib:admfsi}. Apart from the
promising numerical results, the most striking result is that even the simplest
algorithm based on alignment learns recursion.
\end{abstract}

\section{Introduction}

This paper introduces a new type of grammar learning algorithm, which uses the
alignment of sentences to find possible constituents in the form of labelled
brackets. When all possible constituents are found, the algorithm selects the
best constituents. We call this type of algorithm Alignment-Based Learning
(ABL).

The main goal of the algorithm is to automatically find constituents in plain
sentences in an unsupervised way. The only information the algorithm uses stems
from these sentences; no additional information (for example POS-tags) is used.

The underlying idea behind our algorithm is Harris's notion of
interchangeability; \emph{two constituents of the same type can be replaced.}
ABL finds constituents by looking for parts of sentences that can be
replaced and assumes that these parts of the sentences are probably
constituents, which is Harris's notion reversed.

At some point the algorithm may have learned possible constituents that overlap.
Since generating results is done by comparing a learned structure to the
structure in the corpus, the algorithm needs to disambiguate conflicting
constituents. This process continues one tree structure covering the sentence
remains.

This paper is organised as follows. We start out by describing the algorithm in
detail. We then report experimental results from various instances of the
algorithm. We discuss the algorithm in relation to other grammar learning
algorithms, followed by description of some future research.

\section{Algorithm}

In this section we describe an algorithm that learns structure in the form of
labelled brackets on a corpus of natural language sentences. This corpus is a
selection of plain sentences containing no brackets or labels.

The algorithm was developed on several small corpora. These corpora indicated
some problems when simply applying Harris's idea to learn structure. These
problems were solved by introducing two phases: \emph{alignment learning}
and \emph{selection learning}, which will now be described in more detail.

\subsection{Alignment Learning}

The first phase of the algorithm is called \emph{alignment learning}. It 
finds possible constituents by aligning all plain sentences from memory in
pairs. Aligning uncovers parts of the sentences that are similar in both
sentences and parts that are dissimilar. Finally, the dissimilar parts are
stored as possible constituents of the same type. This is shown by grouping
the parts and labelling them with a non-terminal.

Finding constituents like this is based on Harris's notion of
interchangeability. \singleemcite{bib:misl} states that two constituents of
the same type can be replaced. The alignment learning algorithm tries to find
parts of sentences that can be replaced, indicating that these parts might be
constituents.

We have included a simple example taken from the ATIS corpus to give a
visualisation of the algorithm in Table~\ref{tab:bootstrapping}. It shows that
that \textit{Show me} is similar in both sentences and \textit{flights from
Atlanta to Boston} and \textit{the rates for flight 1943} are dissimilar. The
dissimilar parts are then taken as possible constituents of the same type. In
this example there are only two dissimilar parts, but if there were more
dissimilar parts, they would also be grouped. However, a different
non-terminal would be assigned to them (as can be seen in sentences~3 and 4 in
Table~\ref{tab:ambiguous}).

\begin{table}[ht]
\begin{center}
\vskip -.15in
\caption{Bootstrapping structure}
\label{tab:bootstrapping}
\vskip .1in
\begin{tabular}{l}
\textit{Show me flights from Atlanta to Boston}\\
\textit{Show me the rates for flight 1943}\\[.05cm]
\hline
\textit{Show me ( flights from Atlanta to Boston )$_X$}\\
\textit{Show me ( the rates for flight 1943 )$_X$}
\end{tabular}
\end{center}
\end{table}

Note that if the algorithm tries to align two completely dissimilar sentences,
no similar parts can be found at all. This means that no inner structure can be
learned. The only constituents that can be learned are those on sentence
level, since the entire sentences can be seen as dissimilar parts.

\subsubsection{Aligning}

The alignment of two sentences can be accomplished in several ways. Three
different algorithms have been implemented, which will be discussed in more
detail here.

Firstly, we implemented the edit distance algorithm by 
\singleemcite{bib:tstscp} to find the similar word groups in the sentences. It
finds the minimum edit cost to change one sentence into the other based on a
pre-defined cost function $\gamma$. The possible edit operations are insertion,
deletion and substitution, which are used to change one sentence in the other.
It is possible to find the words in the sentences that match (i.e. no edit
operation). These words combined are the similar parts of the two sentences.

The cost function of the edit distance algorithm can be defined to find the
longest common subsequences in two sentences. The cost function $\gamma$ returns
$1$ for an insert or delete operation, $0$ if the two arguments are the same
and $2$ if the two arguments are different. We will call this algorithm the
\emph{default $\gamma$}.

\pagebreak
Unfortunately, this approach has the disadvantage depicted in
Table~\ref{tab:ambiguous}. Here, the algorithm aligns sentences~1 and 2.
Default $\gamma$ finds that \textit{San Francisco} is the longest common
subsequence. This is correct, but results in an unwanted syntactic structure
as can be seen in sentences~3 and 4.

\begin{table}[ht]
\begin{center}
\caption{Ambiguous alignments}
\label{tab:ambiguous}
\vskip .1in
\begin{tabular}{l}
\textbf{1} \textit{from San Francisco to Dallas}\\
\textbf{2} \textit{from Dallas to San Francisco}\\[.05cm]
\hline
\textbf{3} \textit{from ( )$_{X_1}$ San Francisco ( to Dallas )$_{X_2}$}\\
\textbf{4} \textit{from ( Dallas to )$_{X_1}$ San Francisco (
)$_{X_2}$}\\[.05cm]
\hline
\textbf{5} \textit{from ( San Francisco to )$_{X_3}$ Dallas ( )$_{X_4}$}\\
\textbf{6} \textit{from ( )$_{X_3}$ Dallas ( to San Francisco
)$_{X_4}$}\\[.05cm]
\hline
\textbf{7} \textit{from ( San Francisco )$_{X_5}$ to ( Dallas )$_{X_6}$}\\
\textbf{8} \textit{from ( Dallas )$_{X_5}$ to ( San Francisco )$_{X_6}$}
\end{tabular}
\end{center}
\end{table}
 
The problem is that aligning \textit{San Francisco} results in constituents that
differ greatly in length. In other words, the position of \textit{San
Francisco} in both sentences differs significantly. Similarly, aligning
\textit{Dallas} results in unintended constituents (see sentences~5 and
6 in Table~\ref{tab:ambiguous}), but aligning \textit{to} would not (as can
be seen in sentences~7 and 8), since \textit{to} resides more ``in the
middle'' of both sentences.

This problem is solved by redefining the cost function of the edit distance
algorithm to prefer matches between words that have similar offsets in the
sentences. When two words have similar offsets, the cost will be low, but when
the words are far apart, the cost will be higher. We will call this algorithm
\emph{biased $\gamma$}. The biased $\gamma$ is similar to the default $\gamma$,
only in case of a match, the biased $\gamma$ returns
\[
\left|\frac{i_1}{s_1}-\frac{i_2}{s_2}\right|*\frac{s_1+s_2}{2}
\]
where $i_1$ and $i_2$ are the indices of the considered words in sentence~1 and
sentence~2 while $s_1$ and $s_2$ are the lengths of sentence~1 and sentence~2
respectively.

Although biased $\gamma$ solves the problem in Table~\ref{tab:ambiguous}, one
may argue if this solution is always valid. It may be the case that sometimes a
``long distance'' alignment is preferable. Therefore, we implemented a third
algorithm, which does not use the edit distance algorithm. It finds all
possible alignments. In the example of Table~\ref{tab:ambiguous} it finds all
three mutually exclusive alignments.

\subsubsection{Grouping}

The previous section described algorithms that align two sentences and find
parts of the sentences that are similar. The dissimilar parts of the sentences,
i.e. the rest of the sentences, are considered possible constituents. Every
pair of new possible constituents introduces a new non-terminal.\footnote{In
our implementation we used natural numbers to denote the different types.}

\begin{table}[ht]
\begin{center}
\vskip -.15in
\caption{Learning with a partially structured sentence and an unstructured
sentence}
\label{tab:learn1}
\vskip .1in
\begin{tabular}{l}
\textbf{1} What does (AP57 restriction)$_{X_1}$ mean\\
\textbf{2} What does aircraft code D8S mean\\\hline
\textbf{3} What does (AP57 restriction)$_{X_1}$ mean\\
\textbf{4} What does (aircraft code D8S)$_{X_1}$ mean\\
\end{tabular}
\vskip -.1in
\end{center}
\end{table}

At some point the system may find a constituent that was already present in one
of the two sentences. This may occur when a new sentence is compared to a
partially structured sentence in memory. No new type is introduced, instead the
type of the new constituent will be the same type of the constituent in memory.
(See Table~\ref{tab:learn1} for an example.)

\begin{table}[ht]
\begin{center}
\vskip -.15in
\caption{Learning with two partially structured sentences}
\label{tab:learn2}
\vskip .1in
\begin{tabular}{l}
\textbf{1} Explain the (meal code)$_{X_1}$\\
\textbf{2} Explain the (restriction AP)$_{X_2}$\\\hline
\textbf{3} Explain the (meal code)$_{X_3}$\\
\textbf{4} Explain the (restriction AP)$_{X_3}$\\
\end{tabular}
\end{center}
\end{table}

A more complex case may occur when two partially structured sentences are
aligned. This happens when a new sentence that contains some structure, which
was learned in a previous step, is compared to a sentence in memory. When the
alignment of these two sentences yields a constituent that was already present
in both sentences, the types of these constituents are then merged. All
constituents of these types in memory are updated so they have the same type.
This reduces the number of non-terminals in memory as can be seen in
Table~\ref{tab:learn2}.

\subsection{Selection Learning}

The algorithm so far may generate constituents that overlap with other
constituents. In Table~\ref{tab:overlap} sentence~2 receives one structure when
aligned with sentence~1 and a different structure when sentence~3 (which is the
same as sentence~2) is aligned with sentence~4. The constituents in sentence~2
and 3 are overlapping.

\begin{table}[ht]
\begin{center}
\vskip -.1in
\caption{Overlapping constituents}
\label{tab:overlap}
\vskip .1in
\begin{tabular}{l}
\textbf{1} \textit{( Book Delta 128 )$_X$ from Dallas to Boston}\\
\textbf{2} \textit{( Give me all flights )$_X$ from Dallas to Boston}\\\hline
\textbf{3} \textit{Give me ( all flights from Dallas to Boston )$_{Y}$}\\
\textbf{4} \textit{Give me ( help on classes )$_{Y}$}
\end{tabular}
\end{center}
\vskip -.1in
\end{table}

This is solved by adding a selection method that selects constituents until no
overlaps remain. (During the alignment learning phase all possible constituents
are remembered, even if they overlap.) We have implemented three different
methods, although other implementations may be considered. Note that only one of
the methods is used at a time.

\subsubsection{Incremental Method of Constituent Selection}

The first selection method is based on the assumption that once a constituent
is learned and remembered, it is correct. When the algorithm finds a possible
constituent that overlaps with an older constituent, the new constituent is
considered incorrect. We call this method \emph{incr} (after incremental).

The main disadvantage of this method is that once an incorrect constituent has
been learned, it will never be corrected. The incorrect constituent always
remains in memory.

\subsubsection{Probabilistic Methods of Constituent Selection}

To solve the disadvantage of the \emph{incr} method, two additional
(probabilistic) constituent selection methods have been implemented.

The second selection method computes the probability of a constituent counting
the number of times the words in the constituent have occurred as a constituent
in the learned text, normalized by the total number of constituents.
\[P_{leaf}(c)=\frac{|c'\in C:yield(c')=yield(c)|}{|C|}\] where $C$ is the
entire set of constituents. This method is called \emph{leaf} since we count
the number of times the leaves (i.e. the words) of the constituent co-occur in
the corpus as a constituent.

The third method computes the probability of a constituent using the
occurrences of the words in the constituent \emph{and} its non-terminal (i.e.
it is a normalised probability of \emph{leaf}).
\begin{eqnarray*}
\lefteqn{P_{branch}(c|root(c)=r)=}\\
& &\frac{|c'\in C:yield(c')=yield(c)\wedge root(c')=r|}{|c''\in C:root(c'')=r|}
\end{eqnarray*}
The probability is based on the root node and the terminals of the constituent,
which can be seen as a branch (of depth one) in the entire structure of the
sentence, hence the name \emph{branch}.
 
These two methods are probabilistic in nature. The system computes the
probability of the constituent using the formula and then selects constituents
with the highest probability. These methods are accomplished after alignment,
since more specific information (in the form of better counts) can be found at
that time.

\subsubsection{Combination Probability}

Two methods to determine the probability of a constituent have been described.
Since more than two constituents can overlap, a combination of non-overlapping
constituents has to be selected. Therefore, we need to know the probability of
a combination of constituents. The probability of a combination of
constituents is the product of the probabilities of the constituents as in
SCFGs (cf. \npcite{bib:profl}).

Using the product of the probabilities of constituents results in a
\emph{trashing} effect, since the product of probabilities is always smaller
than or equal to the separate probabilities. Instead, we use a normalised
version, the geometric mean\footnote{The geometric mean of a set of
constituents $c_1, \ldots, c_n$ is \(P(c_1\wedge\ldots\wedge
c_n)=\sqrt[n]{\prod_{i=1}^n P(c_i)}\)} \cite{bib:nfomfbfpcp}.

However, the geometric mean does not have a preference for richer structures.
When there are two (or more) constituents that have the same probability, the
constituents have the same probability as their combination and the algorithm
selects one at random.

To let the system prefer more complex structure when there are more
possibilities with the same probability, we implemented the \emph{extended
geometric mean}. The only difference with the (standard) geometric mean is that
when there are more possibilities (single constituents or combinations of
constituents) with the same probability, this system selects the one with the
most constituents. To distinguish between systems that use the geometric mean
and those that use the extended geometric mean, we add a + to the name of the
methods that use the extended geometric mean.

Instead of computing the probabilities of all possible combinations of
constituents, we have used a Viterbi \yrcite{bib:ebfccaaaoda} style algorithm
optimization to efficiently select the best combination of constituents.

\section{Test Environment}

In this section we will describe the systems we have tested and the metrics we
used.

\subsection{System Variables}

The ABL algorithm consists of two phases, alignment learning and selection
learning. For both phases, we have discussed several implementations.

The alignment learning phase builds on the alignment algorithm. We have
implemented three algorithms: \emph{default $\gamma$}, \emph{biased $\gamma$}
and \emph{all} alignments.

After the alignment learning phase, the selection learning phase takes place,
which can be accomplished in different ways: \emph{incr} (the first constituent
is correct), \emph{leaf} (based on the probability of the words in the
constituent) and \emph{branch} (based on the probability of the words
\emph{and} label of the constituent).

There are two ways of combining the probabilities of constituents in the
probabilistic methods: \emph{geometric mean} and \emph{extended geometric
mean}. A + is added to the systems using the extended geometric mean.

The alignment and selection methods can be combined into several ABL systems.
The names of the algorithms are in the form of: \emph{alignment:selection},
where \emph{alignment} and \emph{selection} represent an alignment and
selection method respectively.

\subsection{Metrics}
\label{s:metrics}

To see how well the different systems perform, we use the three following
metrics:
\[NCBP=\frac{\sum_i|O_i|-|Cross(O_i, T_i)|}{\sum_i|O_i|}\]
\[NCBR=\frac{\sum_i|T_i|-|Cross(T_i, O_i)|}{\sum_i|T_i|}\]
\[ZCS=\frac{\sum_i Cross(O_i, T_i)=0}{|TEST|}\]
$Cross(U, V)$ denotes the subset of constituents from $U$ that cross at least
one constituent in $V$. $O_i$ and $T_i$ represent the constituents of a tree in
the learned corpus and in $TEST$, the original corpus, respectively.
\cite{bib:led}

\emph{NCBP} stands for Non-Crossing Brackets Precision, which denotes the
percentage of \emph{learned} constituents that do not overlap with any
constituents in the \emph{original} corpus. \emph{NCBR} is the Non-Crossing
Brackets Recall and shows the percentage of constituents in the \emph{original}
corpus that do not overlap with any constituents in the \emph{learned} corpus.
Finally, \emph{ZCS} stands for 0-Crossing Sentences and represents the
percentage of \emph{sentences} that do not have any overlapping constituents.

\begin{table*}
\caption{Results of the ATIS corpus and OVIS corpus}
\label{tab:results}
\vskip .1in
\begin{center}
\begin{small}
\begin{sc}
\begin{tabular}{|l||r@{\hspace{3pt}}r|r@{\hspace{3pt}}r|r@{\hspace{3pt}}r||
r@{\hspace{3pt}}r|r@{\hspace{3pt}}r|r@{\hspace{3pt}}r|}\hline
\abovespace\belowspace
                   & \multicolumn{6}{c||}{Results ATIS corpus} &
\multicolumn{6}{c|}{Results OVIS corpus} \\\hline
\abovespace\belowspace
&\multicolumn{2}{c|}{NCBP}&\multicolumn{2}{c|}{NCBR}&\multicolumn{2}{c||}{ZCS}&
 \multicolumn{2}{c|}{NCBP}&\multicolumn{2}{c|}{NCBR}&\multicolumn{2}{c|}{ZCS}\\
\hline
\abovespace
default:incr    & 82.55 & (0.80) & 82.98 & (0.78) & 17.15 & (1.17) &
                  88.69 & (1.11) & 83.90 & (1.61) & 45.13 & (4.12)\\
biased:incr     & 82.64 & (0.76) & 83.90 & (0.74) & 17.82 & (1.01) &
                  88.71 & (0.79) & 84.36 & (1.10) & 45.11 & (3.22)\\
all:incr        & 83.55 & (0.63) & 83.21 & (0.64) & 17.04 & (1.19) &
                  89.24 & (1.23) & 84.24 & (1.82) & \textbf{46.84} & (5.02)\\
\hline
default:leaf    & 82.20 & (0.30) & 82.65 & (0.29) & 21.05 & (0.76) &
                  85.70 & (0.01) & 79.96 & (0.02) & 30.87 & (0.07)\\
biased:leaf     & 81.42 & (0.30) & 82.75 & (0.29) & 21.60 & (0.66) &
                  85.32 & (0.02) & 79.96 & (0.03) & 30.87 & (0.09)\\
all:leaf        & 82.55 & (0.31) & 82.11 & (0.32) & 20.63 & (0.70) &
                  85.84 & (0.02) & 79.58 & (0.03) & 30.74 & (0.08)\\
\hline
default:leaf+   & 82.31 & (0.32) & 83.10 & (0.31) & 22.02 & (0.76) &
                  85.67 & (0.02) & 79.95 & (0.03) & 30.90 & (0.08)\\
biased:leaf+    & 81.43 & (0.32) & 83.11 & (0.31) & 22.44 & (0.70) &
                  85.25 & (0.02) & 79.88 & (0.03) & 30.89 & (0.08)\\
all:leaf+       & 82.55 & (0.35) & 82.42 & (0.35) & 21.51 & (0.69) &
                  85.83 & (0.02) & 79.56 & (0.03) & 30.83 & (0.08)\\
\hline
default:branch  & 86.04 & (0.10) & 87.11 & (0.09) & 29.01 & (0.00) &
                  89.39 & (0.00) & 84.90 & (0.00) & 42.05 & (0.02)\\
biased:branch   & 85.31 & (0.11) & \textbf{87.14} & (0.11) & \textbf{29.71} &
(0.00) &
                  89.25 & (0.00) & \textbf{85.04} & (0.01) & 42.20 & (0.01)\\
all:branch      & \textbf{86.47} & (0.08) & 86.78 & (0.08) & 29.57 & (0.00) &
                  \textbf{89.63} & (0.00) & 84.76 & (0.00) & 41.98 & (0.02)\\
\hline
default:branch+ & 86.04 & (0.10) & 87.10 & (0.09) & 29.01 & (0.00) &
                  89.39 & (0.00) & 84.90 & (0.00) & 42.04 & (0.02)\\
biased:branch+  & 85.31 & (0.10) & 87.13 & (0.09) & \textbf{29.71} & (0.00) &
                  89.25 & (0.00) & \textbf{85.04} & (0.00) & 42.19 & (0.01)\\
\belowspace
all:branch+     & \textbf{86.47} & (0.07) & 86.78 & (0.07) & 29.57 & (0.00) &
                  \textbf{89.63} & (0.00) & 84.76 & (0.00) & 41.98 & (0.02)\\
\hline
\end{tabular}
\end{sc}
\end{small}
\end{center}
\end{table*}

\section{Results}

Several ABL algorithms are tested on the ATIS corpus \cite{bib:balacoetpt} and
on the OVIS corpus \cite{bib:admfsi}. The ATIS corpus from the Penn Treebank
is a structured, English corpus and consists of 716 sentences containing 11,777
constituents. The OVIS corpus is a structured, Dutch corpus containing
sentences on travel information. It consists of exactly 10,000 sentences. From
these sentences we have selected all sentences of length larger than one, which
results in 6,797 sentences containing 48,562 constituents.

The sentences of the corpora are stripped of their structure and the ABL
algorithms are applied to them. The resulting structured sentences are then
compared to the structures in the original corpus.

The results of applying the different systems to the ATIS corpus and the OVIS
corpus can be found in Table~\ref{tab:results}. All systems have been tested ten
times, since the \emph{incr} system depends on the order of the sentences and
the probabilistic systems sometimes select constituents at random. The results
in the table show the mean and the standard deviation (in brackets).

\subsection{Evaluation}

Although we argued that the alignment methods \emph{biased} $\gamma$ and
\emph{all} solve problems of the \emph{default} $\gamma$, this can hardly be
seen when looking at the results. The main tendency is that the \emph{all}
methods generate higher precision (NCBP), with a maximum of 89.63~\% on the OVIS
corpus, but that the \emph{biased} $\gamma$ methods result in higher recall
(NCBR) with 87.14~\% on the ATIS corpus and 0-crossing sentences, 29.71~\% on
the ATIS corpus (on the OVIS corpus the maximum is reached with the \emph{all}
method). The \emph{default} $\gamma$ method performs worse overall. These
differences, however, are slight.

The selection learning methods have a larger impact on the differences in the
generated corpora. The \emph{incr} systems perform quite well considering the
fact that they cannot recover from incorrect constituents, with a precision and
recall of roughly 83~\%. The order of the sentences however is quite important,
since the standard deviation of the \emph{incr} systems is quite large
(especially with the ZCS, reaching 1.19~\%).

We expected the probabilistic methods to perform better, but the \emph{leaf}
systems perform slightly worse. The ZCS, however, is somewhat better, resulting
in 22.44~\% for the leaf+ method. Furthermore, the standard deviations of the
\emph{leaf} systems (and of the \emph{branch} systems) are close to 0~\%. The
statistical methods generate more precise results.

The \emph{branch} systems clearly outperform all other systems. Using more
specific statistics generate better results.

The systems using the extended geometric mean result in slightly better results
on the \emph{leaf} system, but when larger corpora are used, this difference
disappears completely.

Although the results of the ATIS corpus and OVIS corpus differ, the conclusions
that can be reached are similar.

\begin{table*}
\begin{center}
\caption{Recursion learned in the ATIS corpus}
\label{tab:recursion}
\vskip .1in
\begin{tabular}{ll}
\textbf{learned} &
\textit{What is the ( name of the ( airport in Boston )$_{18}$ )$_{18}$}\\
\textbf{original} &
\textit{What is ( the name of ( the airport in Boston )$_{NP}$ )$_{NP}$}\\
\textbf{learned} &
\textit{Explain classes QW and ( QX and ( Y )$_{52}$ )$_{52}$}\\
\textbf{original} &
\textit{Explain classes ( ( QW )$_{NP}$ and ( QX )$_{NP}$ and ( Y )$_{NP}$
)$_{NP}$}\\
\end{tabular}
\end{center}
\end{table*}

\subsection{Recursion}

All ABL systems learn recursion on the ATIS and OVIS corpora. Two example
sentences from the ATIS corpus with the original and learned structure can be
found in Table~\ref{tab:recursion}. The sentences in the example are stripped
of all but the interesting constituents to make it easier to see where the
recursion occurs.

The recursion in the first sentence is not entirely the same. The ABL algorithm
finds constituents of some sort of noun phrase, while the constituents in the
ATIS corpus show recursive noun phrases. Likewise in the second sentence, the
ABL algorithm finds a recursive noun phrase while the structure in the ATIS
corpus is similar.

\section{Previous Work}

Existing grammar learning methods can be grouped (like other learning methods)
into supervised and unsupervised methods. Unsupervised methods only use plain
(or pre-tagged) sentences, while supervised methods are first initialised with
structured sentences.

In practice, supervised methods generate better results, since they can adapt
their output to the structured examples from the initialisation phase, whereas
unsupervised methods do not have any idea what the output should look like.
Although unsupervised methods perform worse than supervised methods,
unsupervised methods are necessary for the time-consuming and costly creation
of corpora for which no corpus nor grammar yet exists.

There have been several different approaches to learn syntactic structures. We
will give a short overview here.

Memory based learning (MBL) keeps track of the possible contexts and assigns
word types based on that information \cite{bib:mblaap}.
\singleemcite{bib:pnlumis} describe a method that finds constituent boundaries
using mutual information values of the part of speech n-grams within a
sentence and \singleemcite{bib:diapcfasc} present a method that bootstraps
syntactic categories using distributional information.

\pagebreak
Algorithms that use the minimum description length (MDL) principle build
grammars that describe the input sentences using the minimal number of bits.
This idea stems from the information theory. Examples of these systems can be
found in \singleemcite{bib:giatmdlp} and \singleemcite{bib:ula}.

The system by \singleemcite{bib:ladcag} performs a heuristic search
while creating and merging symbols directed by an evaluation function.
Similarly, \singleemcite{bib:gibhc} describe an algorithm that uses a cost
function that can be used to direct search for a grammar.
\singleemcite{bib:ipgbbmm} describe a more recent grammar induction method
that merges elements of models using a Bayesian framework.
\singleemcite{bib:bgiflm} presents a Bayesian grammar induction method, which is
followed by a post-pass using the inside-outside algorithm
\cite{bib:tgfsr,bib:teoscfgutioa}, while \singleemcite{bib:iorfpbc} apply the
inside-outside algorithm to a partially structured corpus.

The supervised system described by \singleemcite{bib:agiapftatba} takes a
completely different approach. It tries to find transformations that improve a
naive parse, effectively reducing errors.

The two phases of ABL are closely related to some previous work. The
alignment learning phase is effectively a compression technique comparable to
MDL or Bayesian grammar induction methods. However, ABL remembers all possible
constituents, effectively building a search space. The selection learning phase
searches this space, directed by a probabilistic evaluation function. 

It is difficult to compare the results of the ABL system against other systems,
since different corpora or metrics are used. The system described by
\singleemcite{bib:iorfpbc} comes reasonably close to ours. That system learns
structure on plain sentences from the ATIS corpus resulting in 37.35~\%
precision, while the \emph{unsupervised} ABL significantly outperforms this
method, reaching 86.47~\% precision. Only their \emph{supervised} version
results in a slightly higher precision of 90.36~\%.

A system that simply builds right branching structures results in 82.70~\%
precision and 92.91~\% recall on the ATIS corpus, where ABL got 86.47~\% and
87.14~\%. These good results could be expected, since English is a right
branching language; a left branching system performed much worse (32.60~\%
precision and 76.82~\% recall. On a Japanese (a left branching language)
corpus, right branching would not do very well. Since ABL does not have a
preference for direction built in, we expect ABL to perform similarly on a
Japanese corpus compared to the ATIS corpus.

\section{Discussion and Future Extensions}

We will discuss several problems of ABL and suggest possible solutions to
these problems.

\subsection{Wrong Syntactic Type}

There are cases in which the implication ``if two parts of sentences can be
replaced, they are constituents of the same type'', we use in this system,
does not hold. Consider the sentences in Table~\ref{tab:wellmeat}. When
applying the ABL learning algorithm to these sentences, it will determine that
\textit{morning} and \textit{nonstop} are of the same type. However, in the
ATIS corpus, \textit{morning} is tagged as an \textit{NN} (a noun) and
\textit{nonstop} is a \emph{JJ} (an adjective).

\begin{table}[ht]
\begin{center}
\caption{Wrong syntactic type}
\label{tab:wellmeat}
\vskip .1in
\begin{tabular}{l}
\textit{Show me the ( morning )$_X$ flights}\\
\textit{Show me the ( nonstop )$_X$ flights}
\end{tabular}
\end{center}
\end{table}

The constituent \textit{morning} can also be used as a noun in other contexts,
while \textit{nonstop} never will. This information can be found by looking at
the distribution of the contexts of constituents in the rest of the corpus.
Based on that information a correct non-terminal assignment can be made.

\subsection{Weakening Exact Match}

Aligning two dissimilar sentences yields no structure. However, if we weaken
the exact match between words in the alignment phase, it \emph{is} possible to
learn structure even with dissimilar sentences.

Instead of linking exactly matching words, the algorithm should match words
that are equivalent. One way of implementing this is by using \emph{equivalence
classes}. With equivalence classes, words that are closely related are grouped
together. (\singleemcite{bib:diapcfasc} describe an unsupervised way of finding
equivalence classes.)

Words that are in the same equivalence class are said to be sufficiently
equivalent and may be linked. Now sentences that do not have words in common,
but do have words from the same equivalence class in common, can be used to
learn structure.

When using equivalence classes, more constituents are learned since more
terminals in constituents may be seen as similar (according to the equivalence
classes). This results in structures containing more possible constituents from
which the selection phase may choose.

\subsection{Alternative Statistics}

At the moment we have tested two different ways of computing the probability of
a bracket: \emph{leaf} and \emph{branch}. Of course, other systems can be
implemented. One interesting possibility takes a DOP-like approach
\cite{bib:bgaebtol}, which takes into account the inner structure of the
constituents. As can be seen in the results, the system that uses more
specific statistics performs better.

\section{Conclusion}

We have introduced a new grammar learning algorithm based on aligning plain
sentences; neither pre-labelled or bracketed nor pre-tagged sentences are used.
It aligns sentences to find dissimilarities between sentences. The alignments
are not limited to window-size, instead arbitrarily large contexts are used.
The dissimilarities are used to find all possible constituents from which the
algorithm selects the most probable ones afterwards.

Three different alignment methods and five different selection methods have been
implemented. The instances of the algorithm have been applied to two corpora
of different size, the ATIS corpus (716 sentences) and the OVIS corpus (6,797
sentences), generating promising numerical results. Since these corpora are
still relatively small, we plan to apply the algorithm to larger corpora.

The results showed that the different selection methods have a larger impact
than the different alignment methods. The selection method that uses the most
specific statistics performs best. Furthermore, the system has the ability to
learn recursion.

\section*{Acknowledgements}

The author would like to thank Rens Bod and Mila Groot for their suggestions and
comments on this paper, Lo van den Berg for his help generating the results and
three anonymous reviewers for their useful comments on an earlier draft.


\begin{thebibliography}{}

\bibitem[Baker, 1979][Baker][1979]{bib:tgfsr}
Baker, J.~K. (1979).
\newblock Trainable grammars for speech recognition.
\newblock {\em Speech Communication Papers for the Ninety-seventh Meeting of
  the Acoustical Society of America} (pp.\/ 547--550).

\bibitem[Bod, 1998][Bod][1998]{bib:bgaebtol}
Bod, R. (1998).
\newblock {\em Beyond grammar --- an experience-based theory of language}.
\newblock Stanford, CA: {CSLI} Publications.

\bibitem[Bonnema et~al.\/, 1997][Bonnema et~al.\/][1997]{bib:admfsi}
Bonnema, R., Bod, R., \& Scha, R. (1997).
\newblock A {DOP} model for semantic interpretation.
\newblock {\em Proceedings of the {A}ssociation for {C}omputational
  {L}inguistics/{E}uropean {C}hapter of the {A}ssociation for {C}omputational
  {L}inguistics, Madrid} (pp.\/ 159--167).
\newblock Sommerset, NJ: Association for Computational Linguistics.

\bibitem[Booth, 1969][Booth][1969]{bib:profl}
Booth, T. (1969).
\newblock Probabilistic representation of formal languages.
\newblock {\em Conference Record of 1969 Tenth Annual Symposium on Switching
  and Automata Theory} (pp.\/ 74--81).

\bibitem[Brill, 1993][Brill][1993]{bib:agiapftatba}
Brill, E. (1993).
\newblock Automatic grammar induction and parsing free text: A
  transformation-based approach.
\newblock {\em Proceedings of the {A}ssociation for {C}omputational
  {L}inguistics} (pp.\/ 259--265).

\bibitem[Caraballo \& Charniak, 1998][Caraballo and
  Charniak][1998]{bib:nfomfbfpcp}
Caraballo, S.~A., \& Charniak, E. (1998).
\newblock New figures of merit for best-first probabilistic chart parsing.
\newblock {\em Computational Linguistics}, {\em 24}, 275--298.

\bibitem[Chen, 1995][Chen][1995]{bib:bgiflm}
Chen, S.~F. (1995).
\newblock Bayesian grammar induction for language modeling.
\newblock {\em Proceedings of the {A}ssociation for {C}omputational
  {L}inguistics} (pp.\/ 228--235).

\bibitem[Cook et~al.\/, 1976][Cook et~al.\/][1976]{bib:gibhc}
Cook, C.~M., Rosenfeld, A., \& Aronson, A.~R. (1976).
\newblock Grammatical inference by hill climbing.
\newblock {\em Informational Sciences}, {\em 10}, 59--80.

\bibitem[Daelemans, 1995][Daelemans][1995]{bib:mblaap}
Daelemans, W. (1995).
\newblock Memory-based lexical acquisition and processing.
\newblock In P.~Steffens (Ed.), {\em Machine translation and the lexicon}, vol.
  898 of {\em Lecture Notes in Artificial Intelligence},  85--98. Berlin:
  Springer Verlag.

\bibitem[de~Marcken, 1996][de~Marcken][1996]{bib:ula}
de~Marcken, C.~G. (1996).
\newblock {\em Unsupervised language acquisition}.
\newblock Doctoral dissertation, Department of Electrical Engineering and
  Computer Science, Massachusetts Institute of Technology, Cambridge, MA.

\bibitem[Gr{\"u}nwald, 1994][Gr{\"u}nwald][1994]{bib:giatmdlp}
Gr{\"u}nwald, P. (1994).
\newblock A minimum description length approach to grammar inference.
\newblock In G.~Scheler, S.~Wernter and E.~Riloff (Eds.), {\em Connectionist,
  statistical and symbolic approaches to learning for natural language}, vol.
  1004 of {\em Lecture Notes in {AI}},  203--216. Berlin: Springer Verlag.

\bibitem[Harris, 1951][Harris][1951]{bib:misl}
Harris, Z. (1951).
\newblock {\em Methods in structural linguistics}.
\newblock Chicago, IL: University of Chicago Press.

\bibitem[Lari \& Young, 1990][Lari and Young][1990]{bib:teoscfgutioa}
Lari, K., \& Young, S.~J. (1990).
\newblock The estimation of stochastic context-free grammars using the
  inside-outside algorithm.
\newblock {\em Computer Speech and Language}, {\em 4}, 35--56.

\bibitem[Magerman \& Marcus, 1990][Magerman and Marcus][1990]{bib:pnlumis}
Magerman, D., \& Marcus, M. (1990).
\newblock Parsing natural language using mutual information statistics.
\newblock {\em Proceedings of the National Conference on Artificial
  Intelligence} (pp.\/ 984--989).
\newblock Cambridge, MA: MIT Press.

\bibitem[Marcus et~al.\/, 1993][Marcus et~al.\/][1993]{bib:balacoetpt}
Marcus, M., Santorini, B., \& Marcinkiewicz, M. (1993).
\newblock Building a large annotated corpus of english: the {P}enn treebank.
\newblock {\em Computational Linguistics}, {\em 19}, 313--330.

\bibitem[Pereira \& Schabes, 1992][Pereira and Schabes][1992]{bib:iorfpbc}
Pereira, F., \& Schabes, Y. (1992).
\newblock Inside-outside reestimation from partially bracketed corpora.
\newblock {\em Proceedings of the {A}ssociation for {C}omputational
  {L}inguistics} (pp.\/ 128--135).
\newblock Newark, Delaware.

\bibitem[Redington et~al.\/, 1998][Redington et~al.\/][1998]{bib:diapcfasc}
Redington, M., Chater, N., \& Finch, S. (1998).
\newblock Distributional information: A powerful cue for acquiring syntactic
  categories.
\newblock {\em Cognitive Science}, {\em 22}, 425--469.

\bibitem[Sima'an, 1999][Sima'an][1999]{bib:led}
Sima'an, K. (1999).
\newblock {\em Learning efficient disambiguation}.
\newblock Doctoral dissertation, Institute for Language, Logic and Computation,
  Universiteit Utrecht.

\bibitem[Stolcke \& Omohundro, 1994][Stolcke and Omohundro][1994]{bib:ipgbbmm}
Stolcke, A., \& Omohundro, S. (1994).
\newblock Inducing probabilistic grammars by bayesain model merging.
\newblock {\em Second International Conference on Grammar Inference and
  Applications} (pp.\/ 106--118).
\newblock Berlin: Springer Verlag.
\newblock Alicante, Spain.

\bibitem[Viterbi, 1967][Viterbi][1967]{bib:ebfccaaaoda}
Viterbi, A. (1967).
\newblock Error bounds for convolutional codes and an asymptotically optimum
  decoding algorithm.
\newblock {\em Institute of Electrical and Electronics Engineers Transactions
  on Information Theory}, {\em 13}, 260--269.

\bibitem[Wagner \& Fischer, 1974][Wagner and Fischer][1974]{bib:tstscp}
Wagner, R.~A., \& Fischer, M.~J. (1974).
\newblock The string-to-string correction problem.
\newblock {\em Journal of the Association for Computing Machinery}, {\em 21},
  168--173.

\bibitem[Wolff, 1982][Wolff][1982]{bib:ladcag}
Wolff, J.~G. (1982).
\newblock Language acquisition, data compression and generalization.
\newblock {\em Language \& Communication}, {\em 2}, 57--89.

\end{thebibliography}
\bibliographystyle{mlapa}

\end{document}